# Improved Frame Level Features and SVM Supervectors Approach for the Recogniton of Emotional States from Speech: Application to categorical and dimensional states

Imen Trabelsi, Dorra Ben Ayed, Noureddine Ellouze
Université Tunis El Manar
Ecole Nationale d'Ingénieurs de Tunis-ENIT
Laboratoire Signal, Image et Technologies de l'Information-LRSITI
1002, Tunis, Tunisia

*Abstract*— The purpose of speech emotion recognition system is to classify speaker's utterances into different emotional states such as disgust, boredom, sadness, neutral and happiness.

Speech features that are commonly used in speech emotion recognition (SER) rely on global utterance level prosodic features. In our work, we evaluate the impact of frame-level feature extraction. The speech samples are from Berlin emotional database and the features extracted from these utterances are energy, different variant of mel frequency cepstrum coefficients (MFCC), velocity and acceleration features. The idea is to explore the successful approach in the literature of speaker recognition GMM-UBM to handle with emotion identification tasks. In addition, we propose a classification scheme for the labeling of emotions on a continuous dimensional-based approach.

## I. INTRODUCTION

Speech emotion recognition (SER) is an extremely challenging task in the domain of human-robot interfaces and affective computing and has various applications in call centers [1] , intelligent tutoring systems [2], spoken language research [3] and other research areas. The primary channels for robots to recognize human's emotion include facial expressions, gesture and body posture. Among these indicators, the speech is considered as a rapid transfer of complex information. This signal provides a strong interface for communication with computers. Many kind of acoustic features have been explored to build the emotion models [4].

Various classification methods have been verified for emotional pattern classification such as hidden markov models [5], gaussian mixture [6], artificial neural network [7] and support vector machines [8]. In our paper, we investigate the relationship between generative method based GMM and discriminative method based SVM [9, 10]. In addition, we present two approaches, a categorical-based approach, modeling emotions in terms of distinct and discrete categories and a dimensional-based approach, modeling emotions on a continuous space, in which an emotion is mapped within a bipolar dimension: valence and arousal. The valence dimension refers to how positive or negative emotion is. The arousal dimension refers to how excited or not excited emotion is (see fig.3). This concept has gained much attention in recent years.

The rest of paper is organized as follows: First, the description of the proposed speech emotion recognition system. Second, the experimental results of the system. Conclusion is drawn in the final section.

## II. EMOTION RECOGNITION SYSTEM

The proposed speech emotion recognition system contains three main modules (see fig.1) namely (1) extraction of feature, (2) learning the models using machine learning techniques and (3) evaluation of models. First, suitable data sets for training and testing are collected. Second, relevant features are extracted. Third, the extracted features are modeled. Fourth, a set of machine learning techniques could be used to learn the training models. Finally, testing unknown emotional samples are used to evaluate the performances of the models.

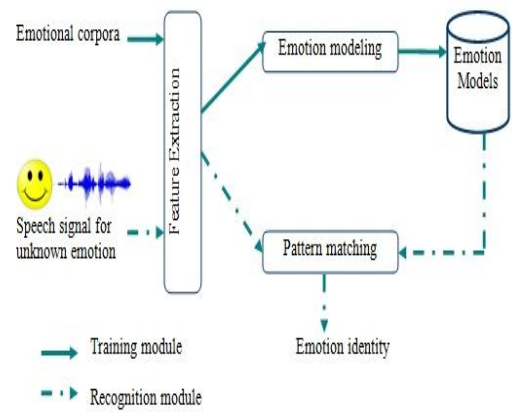

Figure 1. Structure of the speech emotion recognition system

## A. Feature extraction

The first problem that occur when trying to build a recognition framework is the discrimination of the features to be used. Common acoustic features used to build the emotion model include pitch, intensity, voice quality features and formants [9]. Others include cepstral analysis [4]. These features can be divided into two categories: utterance-level features [10] and frame-level features [11].

In this paper, our feature extractor is based on: Mel Frequency Cepstral Coefficients (MFCCs), MFCC-low, energy, velocity and acceleration coefficients. They are extracted on the frame level.

- MFCCs have been the most popular low-level features. They demonstrate good performance in speech and speaker recognition. We use the advantage of this representation for our emotion identification task.

- MFCC-Low are a variant of MFCC. Mel filter banks are placed in [20-300] Hz. Our reason for introducing MFCC-low was to represent pitch variation.

- Energy is an important prosodic feature of speech. It is, often referred to as the volume or intensity of the speech, is also known to contain valuable information [13]. Studies have shown that short term energy has been one of the most important features which provide information that can be used to distinguish between different sets of emotions.

- Velocity (delta) and acceleration (delta-delta) parameters have been shown to play an important role in capturing the temporal characteristics between the different frames that can contribute to a better discrimination [14]. The time derivative is approximated by differentiating between frames after and before the current. It has become common to combine both dynamic features and static features.

## B. The acoustic emotion gaussians model

GMMs have been successfully employed in emotion recognition [15]. The probability density function of the feature space for each emotion is modeled with a weighted mixture of simple gaussian components.

$$P(x) = \sum_{i=1}^{N} W_i N(x; \mu_i, \varepsilon_i). \quad (1)$$

where $N(;,)$ is the gaussian density function, $w_i$, $\mu_i$ and $\varepsilon_i$ are the weight, mean and covariance matrix of the i-th gaussian component, respectively.

This module is assured by the construction of a universal background model (UBM), which is trained over all emotional classes. There are a number of different parameters involved in the UBM training process, which are the mean vector, covariance matrix and the weight.

These parameters are estimated using the iterative expectation-maximization (EM) algorithm [17]. Each emotional utterance is then modeled separately by adapting only the mean vectors of UBM using Maximum A Posteriori (MAP) criterion [18], while the weights and covariance matrix were set to the corresponding parameters of the UBM. To use a whole utterance as a feature vector, we transform the acoustic vector sequence to a single vector of fixed dimension. This vector is called supervector and it takes the form as:

$$\mu = \begin{bmatrix} \mu_1 \\ \mu_2 \\ . \\ . \\ \mu_N \end{bmatrix} \quad (2)$$

This transformation allows the production of features with a fixed dimension for all the utterances. Therefore, we can use the GMM supervectors as input for SVM classifier.

## C. SVM Classification Algorithm

The support vector machines (SVM) [19] are supervised learning machines that find the maximum margin hyperplane separating two classes of data. SVM solve non-linear problems by projecting the input features vectors into a higher dimensional space by means of a mercer kernel.

This powerful tool is explored for discriminating the emotions using GMM mean supervectors. The reason for choosing the SVM classifier for this task is that, it will provide better discrimination even with a high dimension feature space. In our research, we give each training supervector sample with the corresponding emotion class label. After that, we input them to the SVM classifier and gain a SVM emotional model. The output of each model is given to the decision logic. The model having the best score determines the emotion statue. The output of the matching step is a posteriori probability.

In this work, we investigate two SVM kernels in the proposed GMM supervector based SVM: linear and gaussian RBF kernels. The two kernels, take the form as equations (3) and (4) respectively.

$$k(x, v_i) = x.v_i. \quad (3)$$

$$k(x, v_i) = \exp\left[-\frac{1}{2\sigma}(x - v_i)^2\right]. \quad (4)$$

where x is the input data, $v_i$ are the support vectors and $\sigma$ is the width of the radial basis function.

We select in each experiment the best of the two kernels. One against one strategy is used for multi-class classification.

Our experiments are implemented using the LibSvm [20]. The whole speech emotion recognition is shown in Fig. 2.

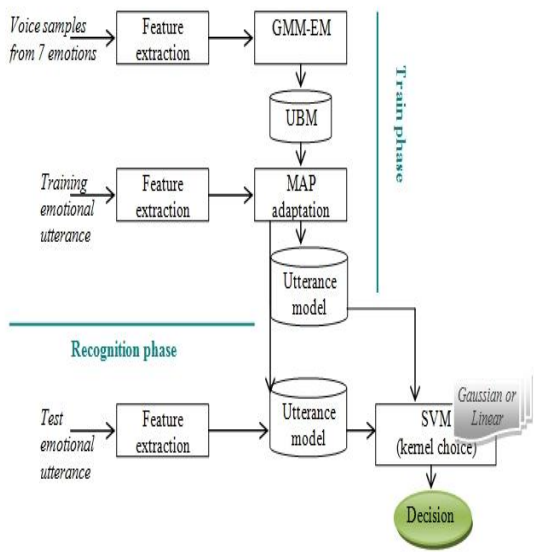

Figure 2. UBM-SVM based speech emotion recognition.

## III. EXPERIMENTS AND RESULTS

### A. Emotional speech database

The database used in this paper is the Berlin database of emotional speech (EMO-DB) which is recorded by speech workgroup leaded in the anechoic chamber of the Technical University in Berlin. It is a simulated open source speech database. This database contains about 500 speech samples proven from ten professional native German actors (5 actors and 5 actresses), to simulate 7 different emotions.

The length of the speech samples varies from 2 seconds to 8 seconds. Table 1 summarizes the different emotions.

TABLE.1 NUMBER OF UTTERANCES BELONGING TO EACH EMOTION CATEGORY

| Emotion | Label | Number |
|---------|-------|--------|
| Anger | A | 128 |
| Boredom | B | 81 |
| Disgust | D | 44 |
| Fear | F | 69 |
| Happiness | H | 71 |
| Sadness | S | 45 |
| Neutral | N | 62 |

Fig. 3 illustrates the distribution of this set of emotions in the two-dimensional space valence and arousal.

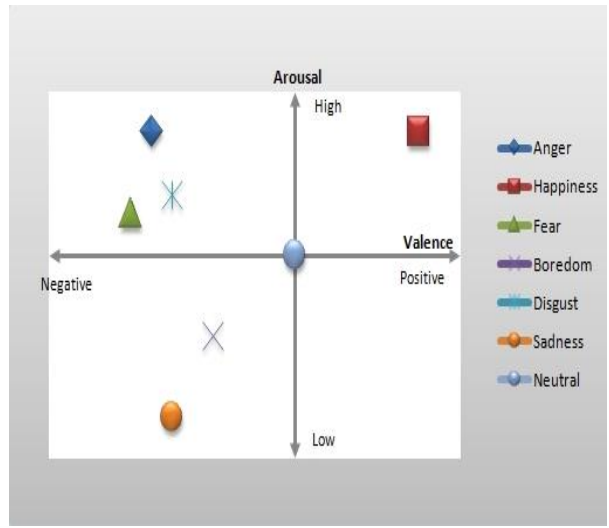

Figure 3. Distribution of the seven emotions in valence-arousal space.

### B. System Description

The data were recorded at a sample rate of 16 KHZ and a resolution of 16 bits. First, the signal is segmented into speech and silence. Then, silence segments are thrown away and the speech segments are pre-emphasized with a coefficient of 0.95. From pre-emphasized speech, each feature vector is extracted from at 8 ms shift using a 16 ms analysis window. A hamming window is applied to each signal frame to reduce signal discontinuity.

Our baseline system is built using 128 UBM gaussian component from the acoustic data of different emotional sentences. Individual emotion models are MAP-adapted. Only the mean vectors are adapted with a relevance factor of 16.

### C. Results and discussion

#### 1) Categorical emotion results

In these experiments, we diverse emotions labeling to discrete states.

Table 2 presents the results conducted on different variants of MFCC in order to extract the most reliable feature.

TABLE.2 RECOGNITION RATE FROM DIFFERENT VARIANT OF MFCC

| Data | Range of filter banks | Recognition rate (%) |
|------|----------------------|----------------------|
| MFCC | 300-3400 | 72.85 |
| Low-MFCC | 0-300 | 62 |
| Combined MFCC | 0-3400 | **81.35** |

Combination of MFCC and MFCC-low led to an accuracy of 81.35%. MFCC-low features perform well in comparison with the small scale of filter banks used, it may be due to its ability to capture voice source quality variations. For the rest of the paper, we choose the combined MFCC (CMFCC).

The table below (table 3) shows the full feature set used for evaluation.

TABLE 3. DIFFERENT SPEECH FEATURE VECTORS

| Data | Features | Size of features |
|---|---|---|
| Data1 | Combined MFCC | 12 |
| Data2 | Combined MFCC+Log Energy | 13 |
| Data3 | Combined MFCC+ Δ | 24 |
| Data4 | Combined MFCC + Δ(MFCC) + Δ Δ(MFCC) | 36 |
| Data5 | Combined MFCC+ Δ(MFCC) + Δ Δ(MFCC) + Δ(log energy)+ Δ Δ(log energy) | 39 |

Table 4 presents the results from a series of recognition experiments to determine the effect of different frame-level features performance. As it can be seen, the recognition rate is varied between (79.50%) and (83.36%). We can conclude from these results that we can get an accuracy of 81.35% with only 12 features comparing with an accuracy of 83.36% with the total 39 features.

TABLE. 4 RECOGNITION RATE BY USING DIFFERENT FEATURES

| Data Feature | Recognition rate(%) |
|---|---|
| Data1 | 81,35 |
| Data2 | 82,12 |
| Data3 | 79,92 |
| Data4 | 79,50 |
| Data5 | 83,36 |

Fig. 4 shows emotion recognition accuracies by analysis over all emotions associated with all previous studied data.

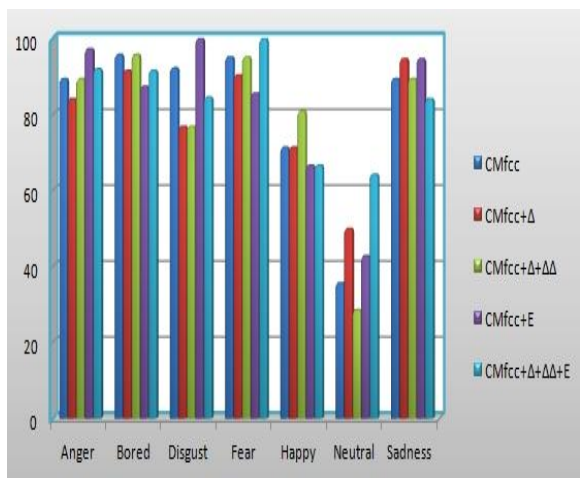

Figure 4. Summary of emotion recognition accuracies over all emotions

We can observe that negative emotion (sadness, boredom, disgust, fear) got the highest classification rate; this could be attributed to the exaggerated expression of emotion by the actors. The lowest rate was for the neutral synthesized speech at 50%, this cloud be explained by the fact that neutral speech doesn't contain specific emotional information.

The addition of energy is beneficial for emotions like anger (from 89.47% to 97.37%), disgust (from 92.31% to 100%) and sadness (from 81.47% to 84.21%). The addition of derivatives significantly improves the recognition rate at the happy emotion (from 71.43 to 80.95). We also conclude that GMM SVM achieves higher recognition rate even when the training data size is small (45 utterances for sadness).

More detailed results in the confusion matrix (table 5), are further shown to analyze the confusion between different emotions associated with MFCC features. The columns show the emotions that the system tried to induce, and the rows are the output recognized emotions.

From these results, we can see that happiness and anger are the most frequently confused emotions. The confusion is also noted between neutral and boredom. This matrix reveals that there are similarities between different categories of emotions that we will try to understand in the rest of the paper.

TABLE. 5 MISCLASSIFICATION BETWEEN 7 DIFFERENT EMOTIONAL STATES

| Recognized As | Ang | Disg | Fear | Happ | Neut | Sad | Bor |
|---|---|---|---|---|---|---|---|
| Ang | 34 | 1 | 0 | 2 | 1 | 0 | 0 |
| Disg | 0 | 12 | 0 | 0 | 0 | 1 | 0 |
| Fear | 0 | 0 | 20 | 1 | 0 | 0 | 0 |
| Happ | 5 | 0 | 1 | 15 | 0 | 0 | 0 |
| Neut | 0 | 1 | 0 | 0 | 5 | 0 | 8 |
| Sad | 0 | 0 | 0 | 0 | 0 | 17 | 2 |
| Bor | 0 | 0 | 0 | 0 | 0 | 1 | 23 |

*2) Dimensional emotion results*

In these experiments, we diverse emotions labeling to binary arousal and valence. The confusion matrix, which can be seen in table 6, illustrates the classifications of the two arousal classes individually (high vs low). The recognition rate is 98.24% for low arousal and 97.84 % for high arousal. As can be seen, high and low emotions are easy classified.

TABLE. 6 CONFUSION MATRIX OF AROUSAL CLASSIFICATION

| Recognized as | High | Low |
|---|---|---|
| High | 91 | 2 |
| Low | 1 | 56 |
| Accuracy(%) | 97,85 | 98,24 |

In valence, there are 3 classes which are positive, neutral and negative. We will classify the affective states into these classes. The obtained recognition rate is 100% for negative, 21.42% for neutral and 57.14% for positive on the valence dimension separately. The worst performance is observed in classifying the neutral state. The table highlights that positive and neutral emotions were confused with negative emotions with the same arousal characteristics.

TABLE. 7 CONFUSION MATRIX OF VALENCE CLASSIFICATION

| Recognized as | Negative | Neutral | Positive |
|---|---|---|---|
| Negative | 94 | 0 | 0 |
| Neutral | 11 | 3 | 0 |
| Positive | 9 | 0 | 12 |
| Accuracy(%) | 100 | 21,42 | 57,14% |

Table 8 shows the percentage of misclassification between negative and positive emotions. The obtained recognition rate is 100% for negative and 38.09 % for positive emotions. The average classification accuracy achieved was 61.4%.

TABLE.8 CONFUSION MATRIX :NEGATIVE VS POSITIVE

| Emotion | Negative | Positive |
|---|---|---|
| Negative | 94 | 0 |
| Positive | 13 | 8 |
| Accuracy(%) | 100 | 38,09 |

Finally, we make a distinction between emotional and neutral speech. As can be seen in table 9, the phrases belonging to neutral state are totally misclassified.

TABLE.9 CONFUSION MATRIX :EMOTIONAL VS NEUTRAL

| Recognized as | Emotional | Neutral |
|---|---|---|
| Emotional | 108 | 7 |
| Neutral | 14 | 0 |
| Accuracy(%) | 93,91 | 0 |

In contrast to arousal, recognition of valence seems to be very challenging, resulting in no more than 60%. Thus it appears that some emotional states share similar acoustic characteristics which make it difficult to discriminate between these emotions.

Positive emotions are poorly recognized, this is due to the fact that happiness which is a positive expression is generally confused with anger which is a negative expression. Given that these two emotions have exactly the same highest rating on the dimension of arousal that suggests that arousal plays an important role in the recognition of emotions. This is one of the reasons why acoustic discriminability on the valence dimension is still problematic: there are no strong discriminative speech features available to discriminate between positive speech and negative speech.

On the other hand, acoustic features are more discriminative between aroused speech (e.g., anger) and not aroused speech (e.g., sadness).

IV. CONCLUSION

Emotional speech recognition is gaining interest due to the widespread applications into various fields.

In our work, this task has been evaluated using frame level features, modeled by GMM-SVM and tested on EMO-DB. Results showed that MFCC, with filter banks placed in [0-3400] extracted at the frame level outperform the traditional MFCC.

Results in emotion recognition experiments are hard to compare, because different database designs are used. Some use elicited speech, whereas others collect spontaneous emotions, some are multi-speaker and others are not. Different basic emotions sets are considered and different data sets are used. Possibly the work presented in [21] is the closest to this one, as acted speech with the same list of emotions. In this paper, Schwenker describe the use of EMO-DB and utilize RASTA-PLP features. Recognition accuracy obtained is 79%.

In addition, we investigated separability on the valence dimension and on the arousal dimension. We found that the arousal dimension seems to be better modeled than the valence dimension.

Recognizing emotions by computer with high recognition accuracy still remains a challenge due to the lack of a full understanding of emotion in human minds. The problem is extremely complicated and thus, the researchers usually deal with acted emotions, just like in our paper. However, in real situations, different individuals show their emotions in a diverse degree and manner. In our future work, we will try to study the performance of the proposed system in a spontaneous emotional database. We will explore the possibilities of integrating other modalities such as manual gestures and facial expression and combine with the result of some other machine learning methods such as KNN, HMM or Random Forest.

*Imen Trabelsi*

I. Trabelsi received a university diploma in computer science in 2009 from the High Institute of Management of Tunis (ISG-Tunisia), the MS degree signal processing in 2011 from the *Institute of Computer Science of Tunis (ISI-Tunisia)*. She is currently working towards the Ph.D. degree in electrical engineering (signal processing) at the *National School of Engineer of Tunis (ENIT)*. Her areas of interests are speech processing, pattern recognition, emotion recognition and speaker recognition.
   E-mail: trabelsi.imen1@gmail.com

*Dorra Ben Ayed Mezghani*

D. Ayed Mezghani received computer science engineering degree in 1995 from the *National School Computer Science (ENSI-Tunisia)*, the MS degree in electrical engineering (signal processing) in 1997 from the *National School of Engineer of Tunis (ENITTunisia)*, the Ph. D. degree in electrical engineering (signal processing) in 2003 from (ENIT-Tunisia).
She is currently an associate professor in the computer science department at the *High Institute of Computer Science of Tunis (ISI-Tunisia)*. Her research interests include fuzzy logic, support vector machines, artificial intelligence, pattern recognition, speech recognition and speaker identification.
E-mail:Dorra.mezghani@isi.rnu.tn,
DorraInsat@yahoo.fr

*Noureddine Ellouze*

N. Ellouze received a Ph.D. degree in 1977 from *l'Institut National Polytechnique* at Paul Sabatier University *(Toulouse-France)*, and Electronic Engineer Diploma from ENSEEIHT in 1968 at the same University. In 1978, Dr. Ellouze joined the Department of Electrical Engineering at the *National School of Engineer of Tunis (ENIT–Tunisia)*, as assistant professor in statistic, electronic, signal processing and computer architecture. In 1990, he became Professor in signal processing digital signal processing and stochastic process. He has also served as director of electrical department at ENIT from 1978 to 1983, general manager and president of the Research Institute on Informatics and Telecommunication IRSIT from 1987-1990, and president of the Institute in 1990-1994. He is now Director of Signal Processing Research Laboratory LSTS at ENIT, and is in charge of Control and Signal Processing Master degree at ENIT. Pr Ellouze is IEEE fellow since 1987; he directed multiple Masters and Thesis and published over 200 scientific papers both in journals and proceedings. He is chief editor of the scientific journal Annales Maghrébines de l'Ingénieur. His research interest include neural networks and fuzzy classification, pattern recognition, signal processing and image processing applied in biomedical, multimedia, and man machine communication.
E-mail:N.Ellouze@enit.rnu.tn